%% file: emnlp2020.tex
\documentclass[11pt,a4paper]{article}

\usepackage[dvipsnames]{xcolor}
\usepackage[hyperref]{emnlp2020}
\usepackage{times}
\usepackage{latexsym}

\usepackage{graphicx}
\usepackage{subfiles}
\usepackage{float}
\usepackage{colortbl} 
\usepackage{booktabs}
\usepackage{multicol}
\usepackage{multirow}
\usepackage{listings}
\usepackage{dsfont}
\usepackage{mdframed}
\usepackage{amssymb}
\usepackage{enumitem}
\usepackage{amsmath}
\usepackage{soul}
\usepackage{mathptmx}
\usepackage{paracol}
\usepackage{xparse} 
\usepackage{fancyhdr}
\usepackage{subcaption}
\usepackage{todonotes}

\newcommand\T{\rule{0pt}{2.6ex}}        
\newcommand\B{\rule[-1.5ex]{0pt}{0pt}}

\definecolor{teal}{RGB}{135,206,235}
\definecolor{fandango}{RGB}{152, 221, 252}
\definecolor{yellowish}{RGB}{146, 157, 207}
\definecolor{greenish}{RGB}{67, 118, 176}
\definecolor{White}{RGB}{255,255,255}

\usepackage{microtype}

\aclfinalcopy % Uncomment this line for the final submission
\title{What does it mean to be language-agnostic? \\
Probing multilingual sentence encoders for typological properties}

\author{Rochelle Choenni\\ ILLC, University of Amsterdam\\  \texttt{r.m.v.k.choenni@uva.nl} \And Ekaterina Shutova\\
  ILLC, University of Amsterdam \\
  \texttt{e.shutova@uva.nl} \\}

\date{}

\begin{document}
\maketitle
\begin{abstract} 
Multilingual sentence encoders have seen much success in cross-lingual model transfer for downstream NLP tasks. Yet, we know relatively little about the properties of individual languages or the general patterns of linguistic variation that they encode. We propose methods for probing sentence representations from state-of-the-art multilingual encoders (LASER, M-BERT, XLM and XLM-R) with respect to a range of typological properties pertaining to lexical, morphological and syntactic structure. In addition, we investigate how this information is distributed across all layers of the models. Our results show interesting differences in encoding linguistic variation associated with different pretraining strategies. 

\end{abstract}

\section{Introduction}

%The development of word representations \citep{mikolov2013distributed, glove} to pre-train models has led to vast improvements on a variety of NLP tasks. This large success has given rise to the field of representation learning, where deep contextualized word and sentence representations are extracted from models instead \citep{peters2018deep}.
%One major drawback to these models, however, is that they are monolingual and require a large amount of training data. Thus, in practice their use is limited to high-resource languages only. 
%While this line of research has met with success, these representations have come at the cost of interpretability. In order to generate universal representations, these models must generalize and share linguistic information across languages. But, we know relatively little about how they do this and to what extent they encode cross-linguistic similarity and variation.

Large-scale pretraining of word representations \citep{glove} and sentence encoders \citep{peters2018deep,devlin2019bert} has led to substantial performance improvements in a variety of NLP tasks. However, due to their data requirements many of these models are limited to high-resource languages only. Aiming to extend the benefits of large-scale pretraining to low-resource languages, much recent research has focused on the development of multilingual word embeddings \citep{ammar2016massively, chen2018unsupervised} and sentence encoders, such as LASER \citep{artetxe2019massively}, Multilingual BERT (M-BERT) \citep{devlin2019bert}, XLM \citep{lample2019cross} and XLM-R(oBERTa) \citep{conneau2019unsupervised}. These encoders are trained to project words and sentences from multiple languages into a shared multilingual semantic space. Their aim is to encode words and sentences irrespective of their source language, such that their meaning can be captured more universally.
These models rely on different types of neural architectures (e.g. recurrent neural networks and Transformers) and training strategies, i.e. using monolingual (M-BERT, XLM-R) or cross-lingual (LASER) training objectives, or a combination thereof (XLM). Whereas models trained with cross-lingual objectives exploit parallel data for supervision, the models that rely on monolingual data are unsupervised.
Having been trained on many languages, both types of models have a wide cross-lingual applicability. 

While this work has met with success, enabling effective model transfer across languages \citep{wu2019beto}, little is known about the linguistic properties of individual languages that such models encode. Nor do we understand to what extent these models capture the patterns of cross-lingual similarity and variation. In this work, we aim to shed light on these questions by probing sentence representations from multilingual encoders for typological properties of languages. We draw inspiration from the field of typological linguistics, which studies and documents structural and semantic variation across languages. In particular, we probe representations from four state-of-the-art multilingual sentence encoders, i.e. LASER, M-BERT, XLM and XLM-R, that exemplify different architectures and pretraining strategies. We investigate (1) the ability of each model to encode and preserve typological properties of languages; (2) where in the model these properties are encoded; and (3) whether the properties that different types of models encode differ in any systematic way. We hypothesize that the type of pretraining tasks influence the linguistic organization within multilingual encoders. 

In line with existing research on interpretation of neural models \citep{linzen2016assessing,belinkov2017neural,conneau2018you,tenney2019you}, we take a probing classification approach. We train a probing classifier to predict typological features of languages from the sentence encodings produced by the models. We use the World Atlas of Language Structures (WALS) database \citep{wals} as a source of typological information and investigate variation along a wide range of linguistic properties, pertaining to lexical, morphological and syntactic structure. We find that (1) all encoders successfully capture information related to word order, negation and pronouns; however, M-BERT and XLM-R outperform LASER and XLM for a number of lexical and morphological properties; (2) typological properties are persistently encoded across layers in M-BERT and XLM-R, but are more localizable in lower layers of LASER and XLM; (3) the incorporation of a cross-lingual training objective contributes to the model learning an interlingua, while the use of monolingual objectives results in a partitioning to language-specific subspaces. These results indicate that there is a negative correlation between the universality of a model and its ability to retain language-specific information, regardless of architecture. Lastly, we tested XLM's generalization to languages unseen during pretraining and find that it is able to capture their typological properties. % of new languages. 

\section{Related work}
Multilingual encoders have been successfully applied to perform  zero-shot cross-lingual transfer in downstream NLP tasks, such as part of speech (POS) tagging and named entity recognition (NER) \citep{van2019comparison}, dependency and constituency parsing \citep{tran2019zero, kim2020multilingual}, %semantic role labelling (SRL), 
text categorization \citep{nozza2020mask}, cross-lingual natural language inference (XNLI) and question answering (XQA) \citep{lauscher2020zero}. Interestingly, models trained in unsupervised monolingual tasks (M-BERT, XLM-R) exhibit competitive performance to those that rely on cross-lingual objectives and parallel data (LASER, XLM). % Yet, the incorporation of cross-lingual objectives remains a popular approach, with \citet{pires2019multilingual} hinting at their vital role for cross-lingual transfer over divergent languages, and
Recently, \citet{huang2019unicoder} introduced Unicoder\footnote{Code or pretrained models are not publicly available.} that relies on 4 cross-lingual tasks. Improving on M-BERT and XLM on XNLI and XQA, the authors claim that the tasks help learn language relationships from more perspectives. This raises the question of whether multilingual encoders capture linguistic and typological properties differently depending on the type of pretraining tasks. 

To investigate this, we use techniques from the rapidly growing line of research on interpretation of neural models \citep{linzen2016assessing, conneau2018you, peters2018dissecting, tenney2019you}, which has been recently extended to the multilingual setting \citep{pires2019multilingual, csahin2019linspector, ravishankar2019word,  ravishankar2019sentence}. \citet{ravishankar2019word, ravishankar2019sentence} study multilingual sentence encoders using probing tasks of \citet{conneau2018you}, e.g probing for universal properties such as sentence length and tree depth, but do not directly probe for typological information. %\citet{ravishankar2019word, ravishankar2019sentence} applied monolingual probing tasks of \citet{conneau2018you}, e.g. probing for word count, to multilingual sentence encoders without adaptation to a multilingual setting. \todo{What do we mean by "without adaptation to cross-lingual setting"? without modelling cross-lingual variation? We need to be more specific here.} \todo{And we need to say what their findings were} 
In a similar vein, \citet{pires2019multilingual} study how M-BERT generalizes across languages by testing zero-shot cross-lingual transfer in traditional downstream tasks. They only briefly touch on typology by testing generalization across typologically diverse languages in POS tagging and NER, and find that cross-lingual transfer is more effective across similar languages. They ascribe this effect to word-piece overlap, arguing that similar success on distant languages might require a cross-lingual objective. On the contrary, \citet{karthikeyan2020cross} show that cross-lingual transfer can also be successful with zero lexical overlap, arguing that M-BERT's cross-lingual effectiveness stems from its ability to recognize language structure and semantics instead. %In this work, we aim to take a closer look at these language structures learnt by multilingual encoders by directly probing the models for linguistic properties. 
In this work, we take a closer look at these emerging language structures by probing the models for typological properties. 

To the best of our knowledge, our approach comes closest to that of \citet{csahin2019linspector}, who probed non-contextualized multilingual \textit{word} representations for linguistic properties such as case marking, gender system and grammatical mood. We considerably expand on this work by proposing methods to probe multilingual \textit{sentence} encoders and investigating a wider range of typological properties pertaining to lexical, morphological and syntactic structure. Since such models are inclined to learn a language identity \citep{wu2019beto}, we also propose a paired language evaluation set-up, evaluating on languages unseen during training. % to prevent the probes from picking up on this signal. Moreover, unsupervised models like M-BERT and XLM-R have offered a solution to side-step the scarcity of parallel resources. Numerous studies report their impressive zero-shot transfer ability on tasks such as POS, NER, dependency parsing, and cross-lingual natural language inference (XNLI) and question answering (XQA) \citep{lauscher2020zero}. Yet, the incorporation of cross-lingual objectives remains a popular approach, with \citet{pires2019multilingual} hinting at their vital role for cross-lingual transfer over divergent languages, and \citet{huang2019unicoder} recently introducing Unicoder that relies on 3 cross-lingual tasks. 

%Working in a monolingual setting, \citet{tenney2019bert} studied how much each layer in BERT contributes to the encoding of linguistic information. Previous work had shown that lower layers of a language model capture local syntax, while higher layers tend to capture more complex semantics \citep{peters2018dissecting, blevins2018deep}. \citet{tenney2019bert} show that the same ordering emerges in BERT, and that syntactic information is more localizable within the model, while information related to semantic tasks is scattered across many layers. We take a similar approach to test where in the model typological information is encoded and whether it is localized or is rather spread across layers.

Previous research on monolingual model probing has shown that lower layers of a language model capture local syntax, while higher layers tend to capture more complex semantics \citep{peters2018dissecting, blevins2018deep}. \citet{tenney2019bert} %studied how much each layer in BERT contributes to the encoding of linguistic information. They 
show that the same ordering emerges in BERT, and that syntactic information is more localizable within the model, while information related to semantic tasks is scattered across many layers. We take a similar approach to test where in the model typological information is encoded and whether it is localized or spread across layers.

\section{Multilingual sentence encoders}
\textbf{LASER} is a BiLSTM encoder trained with an encoder-decoder architecture and a \textbf{cross-lingual} objective --- machine translation (MT). It has $L=5$ layers with a hidden state size of $H=512$. The encoder performs max-pooling over the last hidden states to produce sentence representations $v \in \mathbb{R}^{1024}$. The decoder LSTM is initialized with the sentence representations and trained on the task of generating sentences in a target language. Both the encoder and decoder are shared across all languages, and the input sentences are tokenized based on a joint byte-pair encoding (BPE) vocabulary. We use the pretrained model available for 93 languages. This model leverages parallel data from a combination of text corpora from the Opus website\footnote{http://opus.nlpl.eu/}.

\textbf{M-BERT} is a bidirectional Transformer with $L=12$ and $H=768$, trained on the \textbf{monolingual} Masked Language Modelling (MLM) and Next Sentence Prediction (NSP) tasks. Apart from being trained on the Wikipedia dumps of multiple languages and using a shared WordPiece vocabulary for tokenization, M-BERT is identical to its monolingual counterpart and does not contain a mechanism to explicitly encourage language-agnostic representations. We use the pretrained Multilingual Cased version that supports 104 languages. To obtain fixed-length sentence representations from the transformer, we mean-pool over hidden states.

\textbf{XLM} is a bidirectional Transformer with $L=12$ and $H=1024$. We use the pretrained version that uses BPE vocabulary, BERT's \textbf{monolingual} MLM objective and introduces a new \textbf{cross-lingual} variant on this task, translation language modelling (TLM). In TLM two parallel sentences are concatenated and words in both target and source sentence are masked. This allows the model to leverage information from the context in either language to predict the word, thereby encouraging the alignment of representations in both languages. XLM is trained on the 15 XNLI languages only \citep{conneau2018xnli}, that do not cover all languages used for probing in our work (see Appendix \ref{app:xlm-langs}). This allows us to test its ability to generalize to languages unseen during pretraining, when probing for typological features. 

\textbf{XLM-R} is another encoder with $L=12$ and $H=768$, based on a robustly optimized version of BERT in terms of training regime (RoBerta) \citep{liu2019roberta}. RoBerta is trained with vastly more data and compute power, omits the NSP task and introduces dynamic masking, i.e. masked tokens change with training epochs. The XLM-R variant is trained on 100 languages and introduces the use of a Sentence Piece model (SPM) for tokenization. Unlike XLM, XLM-R does not use the cross-lingual TLM objective, but is only trained on the \textbf{monolingual} MLM task.

\section{Probing for typological information}
\label{sec:TypProbing}

\paragraph{Languages} Restricted by the coverage of WALS, we selected 7 language pairs 
based on their similarity\footnote{Except for XLM, the encoders support all languages used.}: 1.~(Russian, Ukrainian), 2.~(Danish, Swedish), 3.~(Czech, Polish), 4.~(Portuguese, Spanish), 5.~(Hindi, Marathi), 6.~(Macedonian, Bulgarian) and 7.~(Italian, French). These pairs are typologically diverse, cover four language families: Germanic, Indic, Romance and Slavic, and include both high- and low-resource languages from the NLP perspective. For each pair, we use sentences in the first language for training and the second for testing. This prevents the classifier from leveraging information by falling back to language identification. Simultaneously, by choosing related languages, we ensure that similar typological properties are captured in both the train and test set. 

\begin{table}[t]
    \begin{center}
     \small
     \begin{tabular}{p{0.6cm}|p{0.6cm}|p{5.2cm}} 
        \midrule 
        Code  & Type & Feature name \\
        \midrule
        37A& Nom & Definite articles \\
        38A& Nom & Indefinite articles\\
        45A& Nom & Politeness distinctions in pronouns\\
        47A & Nom & Intensifiers and reflexive pronouns \\
        51A & Nom &Position of case affixes  \\
        %\B \hline
        70A &Verb& The morphological imperative \\%\T\\
        71A &Verb& The prohibitive\\
        72A &Verb& Imperative-hortative systems\\ 
        79A &Verb& Suppletion according to tense and aspect \\
        79B &Verb& Suppletion in impertatives and hortatives \\
        %\B \hline \T
        81A & WO & Order of Subject, Object and Verb (SOV)\\
        82A & WO & Order of Subject and Verb (SV) \\
        83A & WO & Order of Object and Verb (OV)\\
        85A & WO & Order of adposition and noun phrase\\
        86A & WO & Order of genitive and noun \\
        87A & WO & Order of adjective and noun\\
        92A & WO & Position of polar question particles \\
        93A & WO & Position of interrogative phrases in content questions\\
        95A & WO & Relationship between OV and adposition and noun phrase order\\
        %\B 
        97A & WO &Relationship between OV and adjective and noun order \\
       % \hline \T
         115A & SC & Negative indefinite pronouns and predicate negation\\
        116A & SC & Polar questions \\%\B \\
        %\hline 
        143F & WO &Postverbal negative morphemes \\% \T \\
        144D & WO & Position of negative morphemes\\ %in SVO languages \\
        \B 144J & WO &Subject verb negative word object order\\
        \midrule
      \end{tabular}
      \vspace{-0.3cm}
      \caption{The 25 WALS features used for probing with their correpsonding WALS codes and categories.} 
        \label{table:features}
    \end{center}
    \vspace{-0.1cm}
\end{table}

\paragraph{Typlogical features}  
We extract typological properties of languages from WALS, a large publicly available database consisting of 192 linguistic features annotated by typology experts for 2679 languages. In WALS, each feature is listed with languages and their feature values, e.g. feature \texttt{47A}: `\textit{Intensifiers and Reflexive Pronouns}' $-$ French::`\textit{differentiated}' and Swedish::`\textit{identical}'. Despite its coverage, WALS is quite sparse as only few languages have annotations for all features.  Thus, we selected features containing annotations for at least 4 of our languages and discarded features for which the chosen languages did not show typological diversity. Moreover, we made sure that all feature values were covered by the 15 languages that XLM is trained on. As a result, we probe for 25 features classified by WALS under the categories: Word order (WO), Nominal (Nom) and Verbal (Verb) categories and Simple clauses (SC), each in a separate probing task $\tau_i$ (see Table \ref{table:features}). 

\paragraph{Probing task setup} Per language, we retrieved 10K random sentences from the Tatoeba corpora\footnote{https://tatoeba.org}. We filtered out translations between train and test languages to prevent the classifier from overfitting on semantic meaning. %While we did not explicitly control for sequence length, upon inspection of the resulting datasets, we found that they are naturally distributed similarly with on average 48 characters per sentence\todo{hmm sounds shortish -- let's discuss on Friday}.  %Moreover, as  languages require different numbers of characters to express similar meanings, it did not seem feasible to explicitly control for the distribution in sequence length. Yet, we performed some analysis to inspect the in-between language variation and found that the datasets are naturally distributed similarly with on average 48 characters per sentence.
%Given a set of $N$ input sentences $U_i=[u_0,..,u_n]$ per language $i \in S$, a dataset (X, y) for each of the 25 tasks is created by annotating all sentences from a language with their corresponding feature value $v^{\tau}_{i}$ in WALS: $X_{\tau}=[U_{0},.., U_{|S|}] \mapsto y_{\tau}=[v^{\tau}_{0}\times N,..,v^{\tau}_{|S|} \times N]$. 
Given a set of input sentences per language, a dataset for each of the 25 tasks is created by annotating all sentences from a language with their corresponding feature value in WALS. 
Thus, annotations are at the language-level. % \footnote{This would impose new scalability issues as low resource languages lack high quality annotation tools to correctly detect linguistic phenomena. }
This probing task design relies on the assumption that encoders capture typological properties for any sentence in a language, irrespective of semantic meaning. Following \citet{pires2019multilingual} and \citet{libovicky2019language} we hypothesize that sentence representations from our encoders contain a language-specific component that is similar across all sentences in a language. Table \ref{table:variation} provides an indication of the variation of feature values represented in our dataset. Note that paired languages do not always have the same value for the same typological feature, thus the respective probing tasks would not be possible to solve by falling back to a similar language identification task.

\paragraph{Probing classifier} For probing we use a one-layer MLP with 100 hidden units, ReLU activation, and an output layer that uses the softmax function to predict class labels. The simplicity of the architecture was chosen to limit task-specific training, such that the classifier is forced to rely on information contained in the encoder representations as much as possible. We experimented with various similar architectures and hyperparameter values, but no prominent differences were observed\footnote{Other works used more expressive models, e.g. 300 hidden units \citep{csahin2019linspector} and two-layer MLPs \citep{tenney2019you}. This did not yield substantial changes in our experiments. We report results from the least expressive model tested, as high performance of M-BERT indicates that this model is in principle capable of learning the task, given an informed encoder.}. We freeze the parameters of the sentence encoder during training such that all learning can be ascribed to the probing classifier $P_{\tau}$. The classifier then predicts the feature values $y_{\tau}$ from the representations of the input sentences.
\input{table_variation.tex}

To keep results across different tasks comparable, we perform no fine-tuning on the hyperparameters. For all tasks we train for 20 epochs with early stopping (patience=5), using the Adam optimizer \citep{kingma2014adam}. We set the batch size to 32 and use dropout (rate=0.5). As some features can take $n>2$ values, we encode the labels as one-hot vectors and obtain the non-binary predictions at test time by returning the class with the highest probability. To account for class imbalances, we report results using macro-averaged-F1 scores.

\section{Top-layer probing experiments}\label{sec:top-probing}

We first probe the top-layer representations produced by our encoders, which are commonly used in downstream tasks.

\paragraph{Baseline} To test to what extent the classifier relies on information from the encoder as opposed to information learned from task-specific training, we use randomized encoders as a baseline for comparison. Following \citet{tenney2019you}, we randomized the weight matrices of our pretrained models. We found that our simple classifier is unable to learn from these representations, falling back to majority class voting in all cases. Thus, the performance for all randomized encoders is identical and we report these scores under \textit{Baseline}. 

\begin{table}[t!]
\begin{center}
\small
\begin{tabular}{c|c|cccc|c}
\midrule 
Code  & Type & LASER & M-BERT & XLM & XLM-R & Baseline \\
\midrule
37A & Nom &0.864 & 0.957 & 0.83 & \textbf{0.997} & 0.199\\
38A* & Nom & 0.571 &\textbf{0.597} & 0.595 & 0.579 & 0.334\\
45A$^{\dagger}$ & Nom & 0.997& \textbf{1.0}& 0.989 & \textbf{1.0} & 0.428\\
47A$^{\dagger}$ & Nom & 0.97 & 0.995 & 0.934 & \textbf{0.999}&  0.333\\
51A$^{\ddagger}$ & Nom & 0.682 & \textbf{0.763} & 0.752 & 0.762 & 0.375\\
\midrule
70A & Verb &0.64 &0.69 & 0.603 & \textbf{0.695} &  0.243 \\
71A & Verb & 0.347 & 0.522 & 0.452 & \textbf{0.576} & 0.243\\ 
72A & Verb &0.422 & 0.763& 0.557 & 0.\textbf{769} & 0.417\\
79A$^{\mathsection}$ & Verb &0.456 & 0.94& 0.646 & \textbf{0.978} & 0.4\\
79B$^{\mathsection}$ & Verb &0.212&0.528& 0.382 & \textbf{0.544} &0.25\\
\midrule
81A & WO &0.993&\textbf{ 1.0}& 0.959 & 0.998 & 0.462\\
82A & WO &0.429	& 0.352&\textbf{ 0.449} & 0.368 & 0.363\\
83A & WO & 0.993& \textbf{1.0}&0.939 & 0.999 & 0.462\\
85A & WO & 0.993 & \textbf{1.0}& 0.873 & 0.995 & 0.462\\
86A$^{\dagger}$ & WO & 0.763 & 0.811 & 0.757 & \textbf{0.82}& 0.166\\
87A & WO &0.976 & \textbf{0.999}& 0.944 & 0.998 & 0.416\\
92A$^{\mid}$& WO &0.212 & 0.16& 0.231 & 0.206 & \textbf{0.285}\\
93A$^{\mathparagraph}$ &WO &0.647&0.65& 0.627 & \textbf{0.665} & 0.25\\
95A & WO &0.993& \textbf{1.0}& 0.96 & 0.999 & 0.462\\
97A &WO & 0.983 & 0.996& 0.941 & \textbf{0.998} & 0.243\\
\midrule
115A$^{\#}$ & SC& 0.998 & \textbf{1.0}& 0.984 & 0.999 & 0.4\\
116A$^{\lozenge}$ & SC&0.584&0.622& 0.602 & \textbf{0.634} & 0.4\\
\midrule
143F & WO &0.608 &0.644& 0.599 & \textbf{0.65} & 0.364\\ 
144D$^{\downarrow}$& WO & 0.978 & 0.998& 0.979 & \textbf{1.0} & 0.429\\ 
144J$^{\delta}$ & WO &0.983	& 0.996 & 0.954 & \textbf{0.999} & 0.445\\ 
\midrule
\end{tabular}
\vspace{-0.1cm}
\caption{Macro-averaged-F1 scores on the test set per typological feature. Unless indicated otherwise, all language pairs were used.
Excluded pairs: *:(1), $^{\dagger}$:(1, 3 and 6), 
$^{\ddagger}$:(6 and 7), $^{\mathsection}$:( 2, 4, 5 and 7), $^{\mid}$:(5 and 6), $^{\mathparagraph}$:(1, 4, 6, 7), $^{\#}$:(1-3 and 6), $^{\lozenge}$:(7), $^{\downarrow}$:(3, 5 and 7), $^{\delta}$:(5 and 7).}
\label{sentence-level-results}
\end{center}
\end{table}

\paragraph{Results} In Table \ref{sentence-level-results}, we report the performance over all languages per task. Note that, due to missing values, not all languages were used for each task, as indicated in the table. The results show that all encoders are able to capture features related to word order (e.g. \texttt{81A}, \texttt{85A}, \texttt{95A} and \texttt{97A}), pronouns (\texttt{45A}, \texttt{47A}) and negation (\texttt{144D}, \texttt{144J}). M-BERT and XLM-R, however, generally outperform LASER and XLM when it comes to lexical and morphological properties, such as in the nominal (e.g. \texttt{37A}, \texttt{51A}) and verbal (e.g. \texttt{70-72A}) category tasks. The strongest difference between encoders is observed when probing for the suppletion features (\texttt{79A,B}). Furthermore, for none of the encoders, the classifier is capable of accurately predicting properties related to the form of questions (\texttt{92A}, \texttt{93A}, \texttt{116A}). 
Lastly, we find that, while obtaining a high performance for other word order tasks, the classifier fails to predict the feature \texttt{82A} (\textit{SV order}). %\todo{Perhaps double check that these all still hold after transition to F1 etc. :-)}

%For example, for feature \texttt{70A} (\textit{Morphological imperative}), the classifier fails to predict the label `\textit{Second singular and second plural}' for the Spanish representations from both encoders. Yet, for Polish it is able to predict that same label for both LASER and M-BERT. %($\pm$90-100 \%).However, when checking for specific languages or language families %\footnote{including both high- and low-resource languages}that either LASER or M-BERT consistently fails for, we did not find strong patterns and instead the low performance tends to be associated with specific typological features.

\paragraph{Analysis}
To further analyze our models, we investigated the accuracy per feature broken down by language and specific feature values. Overall, the classifier consistently fails to predict certain features for specific languages, resulting in the per language performance often being either very high or low (see Figure \ref{fig:per_lang_heatmaps}, for XLM and XLM-R see Appendix \ref{sec:xlm-resutls}). This indicates that encoders indeed capture typological properties of languages, irrespective of the sentence. Yet, no languages or language families were found that an encoder always fails for. Instead, low performance tends to be associated with specific features. In addition, XLM obtains performance levels similar to LASER for languages it was not pretrained on. In fact, we found no relationship between the support of language and performance, indicating that XLM successfully generalizes to unseen languages (see Appendix \ref{app:generalization}). 

\begin{figure}[t!]
    \centering
    \vspace{-0.3cm}
    \includegraphics[width=\linewidth, height=14.5cm]{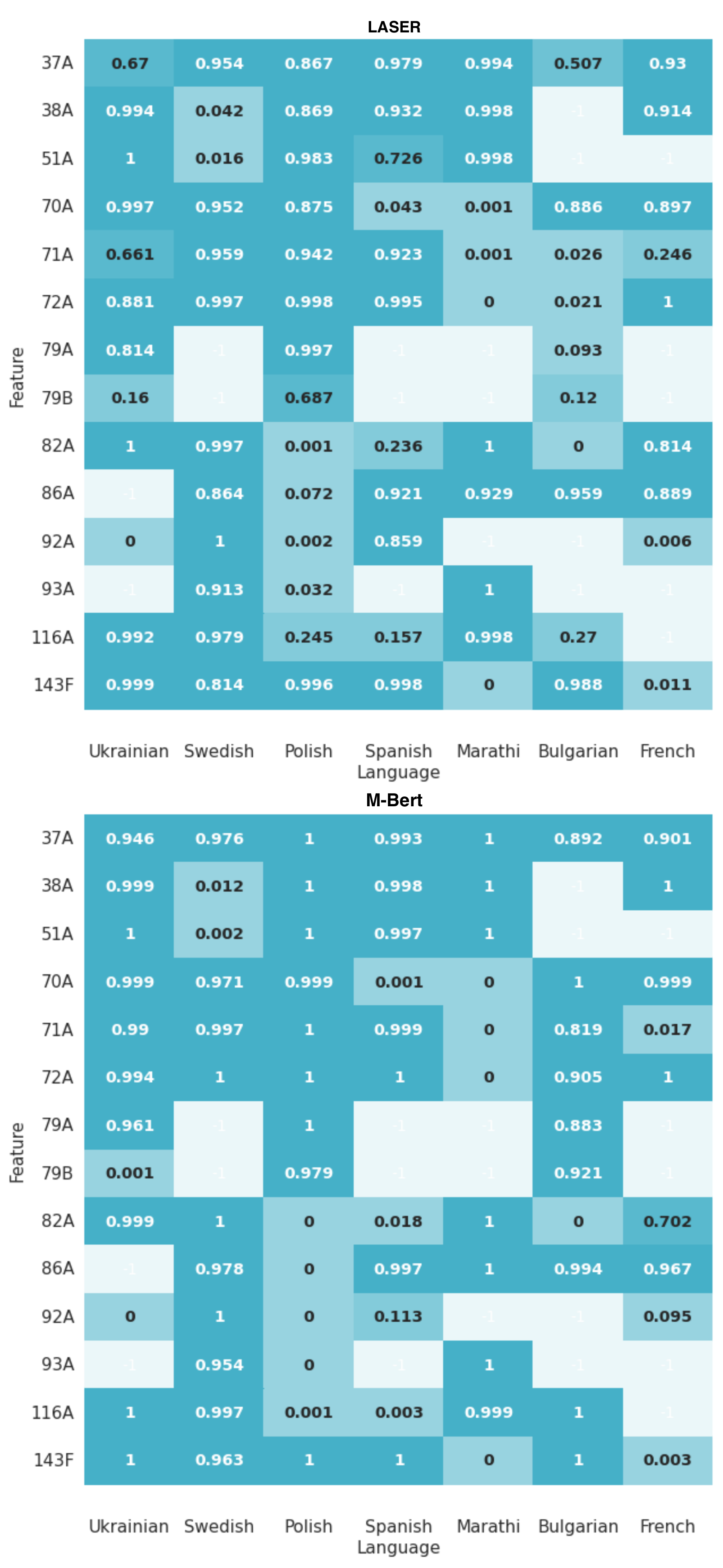}
    \vspace{-0.8cm}
    \caption{Heatmaps of the performance (in \% accuracy) for a selected number of interesting tasks from LASER and M-BERT broken down per language.}
    \label{fig:per_lang_heatmaps}
    \vspace{-0.1cm}
\end{figure}

Moreover, we observe that the classifier may fail both in cases where the labels for the paired test and train languages are identical and in cases where they are different. For instance, despite Bulgarian having the same label as Macedonian, the classifier based on LASER or XLM fails for Bulgarian in multiple tasks (e.g., \texttt{71A}, \texttt{72A}, \texttt{116A}) \footnote{Note that all encoders were trained on Bulgarian.}. On the other hand, there are also cases where the classifier succeeds despite the test language and its most similar training language having a different label, e.g.  LASER for Spanish (\texttt{92A}) and XLM for French (\texttt{70A}). This demonstrates that the classifier does not merely rely on similar language identification.

\begin{figure*}[h!]
    \includegraphics[width=1.05\textwidth, height=4.3cm]{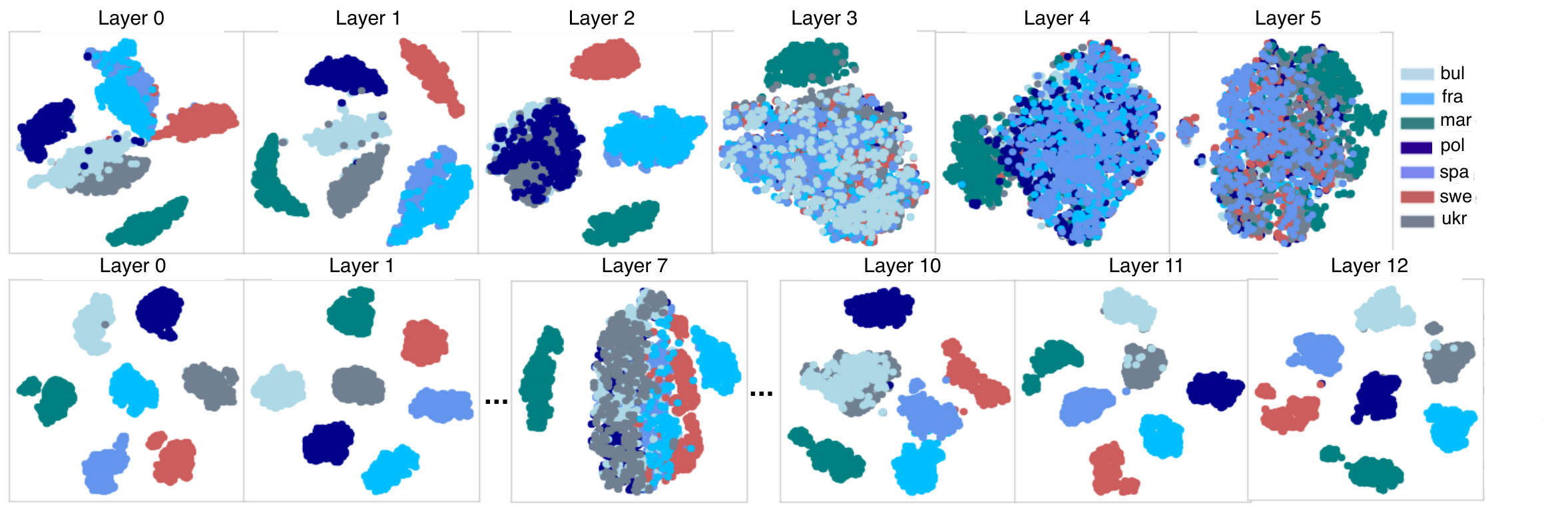}
    \caption{t-SNE plots of representations from layers of LASER (top) and M-BERT (bottom), where layer 0 corresponds to the non-contextualized token embeddings (visualizations made using PCA with k = 10).}
    \label{fig:tsne-laser-layers}
    \vspace{-0.1cm}
\end{figure*}

\paragraph{Languages and feature values}

When comparing the per language performance across encoders, we see that, although LASER and XLM exhibit a lower performance in more languages, there are specific cases in which all encoders fail. Consequently, certain properties of particular languages may be more difficult for encoders to capture. Such cases include, for instance, features `\textit{Indefinite articles}' \texttt{(38A)} and `\textit{Postverbal negative morpheme} \texttt{(143F)} for Swedish and French respectively.
%Such cases include, Swedish, Polish and French for features \texttt{38A}, \texttt{93A} and \texttt{143F}.

In addition, we analysed specific feature values that the encoders fail to capture. 
For example, we find that LASER and XLM fail to predict the label `\textit{Maximal system}' (assigned to Bulgarian and Marathi) for feature \texttt{72A} (\textit{Imperative-hortative systems}). M-BERT and XLM-R, while failing for Marathi, obtain $\pm 90\%$ accuracy for Bulgarian. A similar effect for LASER is observed for other labels of verbal and nominal category tasks, e.g. `\textit{Tense}' (\texttt{79A}: \textit{Suppletion according to tense and aspect}), and to a lesser extent also XLM, e.g. `\textit{Special imperative + special negative}' (\texttt{71A}: \textit{The prohibitive}). No such cases were identified for M-BERT or XLM-R. This observation clarifies our finding that M-BERT and XLM-R outperform LASER and XLM on the majority of nominal and verbal category tasks. Whereas LASER and XLM suffer from both the inability to capture certain feature values as well as specific language-feature combinations, M-BERT and XLM-R only suffer from the latter.  

In the particular case of feature \texttt{82A} (\textit{SV order}), the classifier always fails to predict `\textit{No dominant order}', assigned to Bulgarian, Spanish and Polish, for all encoders. We speculate that this may be due to the fact that encoders are inclined to assign the order predominantly seen during training, without quantifying an extent, thereby forcing an order to non-dominant order languages. Similarly, in the few cases in which $\pm 50\%$ accuracy is obtained, LASER has difficulties specifically with encoding a lack of certain properties, e.g. Ukrainian: `\textit{No definite or indefinite article}' (\texttt{37A}), Spanish: `\textit{No case affixes or adpositional clitics}' (\texttt{51A}), Polish: `\textit{No suppletive imperatives}' (\texttt{79B}).

In order to test the validity of language-level tasks, we repeated experiments for properties specific to questions on a subset of our data, where only questions were used as input sentences. This resulted in a subset of $\approx 10\%$ of the full dataset per language and we obtained similar classifier performance for the features of interest across encoders.

%Lastly, it is worth noting that LASER and XLM only beat M-BERT for feature \texttt{92A} (\textit{Position of Polar Question Particles}). While exhibiting baseline-level performance, LASER correctly distinguishes languages labelled `\textit{No question particle}' (incl. Spanish), from the majority labelled `\textit{Initial}' (incl. Portuguese). Yet, M-BERT obtains low performance across all languages and feature values.

\section{Probing and analysis across layers}

\subsection{Methods}

\paragraph{Layer probing} In the previous experiments, we probed sentence representations from the top layer $H^{(L)}$ of the model. However, each of our models produces a set of activations at each layer: $H^{(0)}, H^{(1)},.., H^{(L)}$, where $H^{(L)}=[\textbf{h}^{(L)}_{t_0},...,\textbf{h}^{(L)}_{t_n}]$ and $n$ is the number of tokens.  To test where in the model the typological properties are encoded, we now probe sentence representations from each layer of the model. We compute per-layer sentence representations by mean-pooling over the corresponding hidden states. We then take the same approach as in Section~\ref{sec:TypProbing} to probe for typological properties. 

\paragraph{Full model probing}
As the layer-wise approach does not take into account the interactions between different layers, we also adapt the method proposed by \citet{tenney2019bert} that borrows the scalar mixing technique from ELMo \citep{peters2018deep}. For each probing task we introduce a set of scalar parameters: $\lambda_\tau$ and $a^{(1)}_{\tau},a^{(2)}_{\tau}, .., a^{(L)}_{\tau} $. We compute per-layer sentence representations by mean-pooling over hidden states $t_0,..,t_n$ as before: $\textbf{h}^{(l)}_{\tau} \text{ = pool}([\textbf{h}^{(l)}_{t_0}, \textbf{h}^{(l)}_{t_1},.., \textbf{h}^{(l)}_{t_n}]), \text{where } \textbf{h}^{(l)}_{t_i} = \overrightarrow{\textbf{h}^{(l)}_{t_i}} + \overleftarrow{\textbf{h}^{(l)}_{t_i}}\text{ for LASER.}$ To pool across layers we use the mixing weights:

\vspace{-0.3cm}
\begin{equation}
\begin{aligned}
        \textbf{h}_{\tau} & = \lambda_\tau \sum^{L}_{l=1} s^{(l)}_\tau \textbf{h}^{(l)}_{\tau}
\end{aligned}
\vspace{-0.2cm}
\end{equation}
\noindent where $s_{\tau}= \text{softmax}(a_\tau)$. These weights $a_\tau$ are jointly learned with each task to give the probing classifier $P_\tau$ access to the full model. Note that we exclude layer 0 as token embeddings in LASER have a different dimensionality from higher layers. After training, we extract the learned coefficients from the probing classifier to estimate the contribution of different layers to the particular task. Higher weights are interpreted as evidence that the corresponding layer contains more information about the typological property. We report the Kullback-Leibler divergence $K(s_\tau) = KL(s_\tau|| \text{Uniform})$\footnote{$KL(p||q) = \sum^{N}_{i=0} p(x_i)\text{log}(\frac{p(x_i)}{q(x_i)})$} for each task as an estimation of the non-uniformity of the statistics. We interpret a higher KL divergence as an indication of a more localizable feature.

\subsection{Results and analysis}

\begin{figure}[t!]
    \centering
  \includegraphics[width=0.8\linewidth, height=7cm]{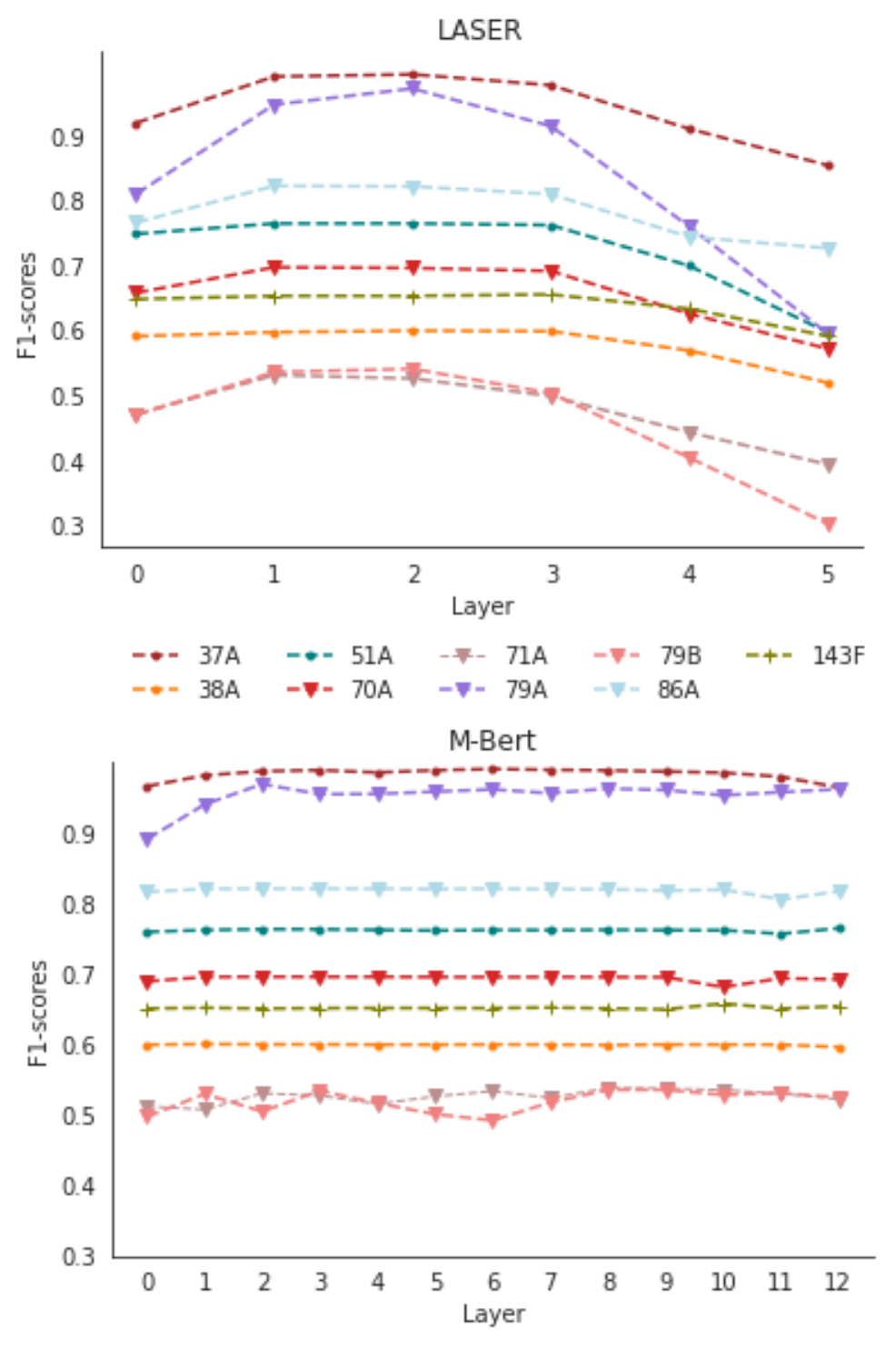}
      \vspace{-0.2cm}
    \caption{macro-averaged-F1 scores for per-layer probing of LASER and M-BERT. }
    \label{fig:per-layer-results}
\end{figure}

Figure \ref{fig:per-layer-results} shows the classifier performance when probing the different layers of the models (see appendix \ref{sec:xlm-resutls} for XLM and XLM-R). We find that in both models that incorporate a cross-lingual objective (LASER and XLM), the typological properties are more prevalent in lower layers of the network (1-2) and performance steadily decreases in higher layers (3+). In contrast, in M-BERT and XLM-R the performance is stable throughout all layers.
 
Figure \ref{fig:mixing_weights} shows the distribution of the learned mixing weights across layers (see Appendix \ref{sec:scalarweights} for XLM and XLM-R). We find that for LASER and XLM the probes almost exclusively rely on information from the first layers, which is in line with our findings from the per-layer results. Given the low KL divergences across tasks, the learned weights remain more uniform for M-BERT and XLM-R. Nevertheless, we observe a trend that middle layers gradually decrease in importance, while the last few layers regain it again. %A faint signal of this trend is found in the layer results of these models, but the performance difference is often negligibly small ($\pm 0.05$). 

These results indicate that in models pretrained with a cross-lingual objective --- LASER and XLM --- typological information is localizable in the lower layers, but is lost in higher layers. In M-BERT and XLM-R, which rely on monolingual pretraining objectives, this information is either captured in the lower layers and correctly propagated through the higher layers, or it could be spread across the model instead.

\begin{figure}[ht]
    \centering
    \includegraphics[width=\linewidth, height=6cm]{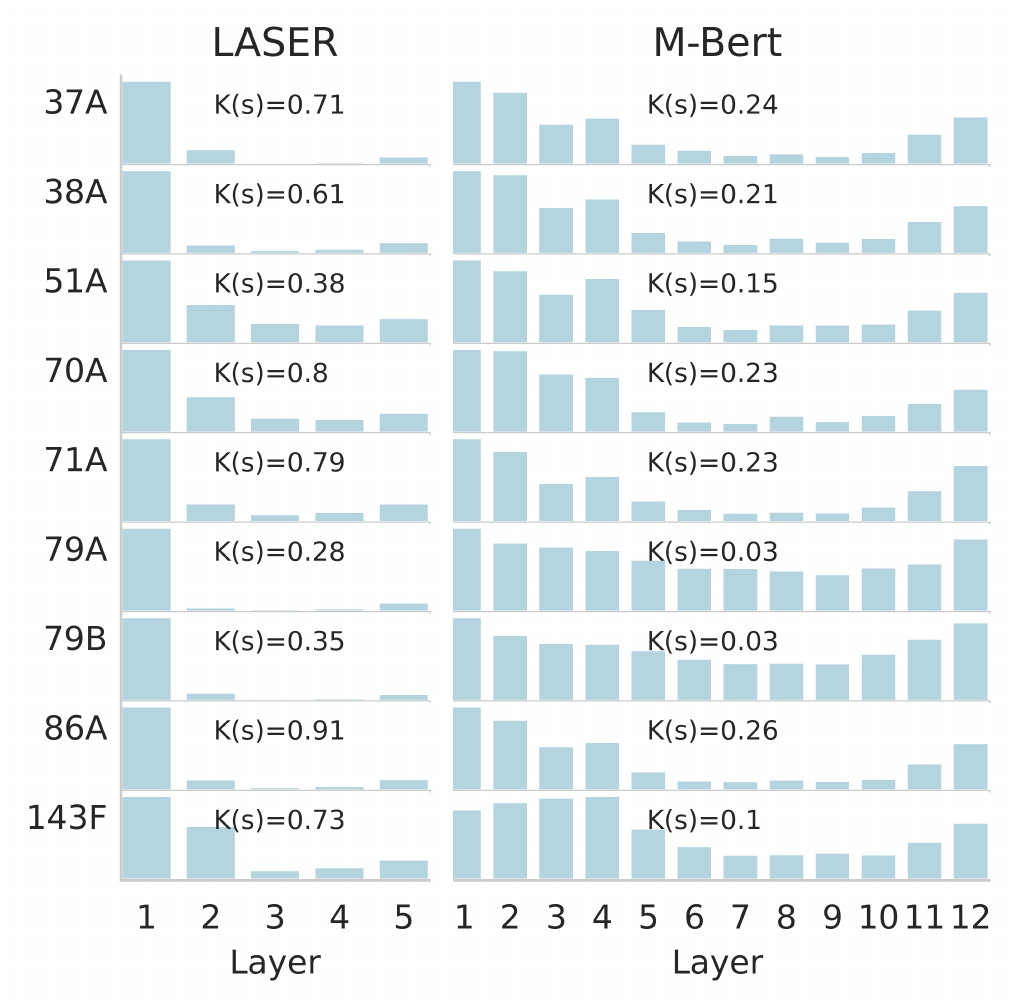}
    \vspace{-0.6cm}
    \caption{Learned mixing weights $s_\tau$ and corresponding KL divergences $K(s)$ for LASER and M-BERT.}
    \label{fig:mixing_weights}
\end{figure}

\paragraph{Universality vs. language-specific information} Previous research suggests that M-BERT partitions its multilingual semantic space into separate language-specific subspaces, and is thus not a true interlingua \citep{libovicky2019language, singh2019bert}. In Figure \ref{fig:tsne-laser-layers} we visualize the representations of all sentences in our test datasets from layers in LASER and M-BERT in a t-SNE plot. In agreement with previous research, we find that in M-BERT and XLM-R languages continue to occupy separate subspaces in the last layer (see Appendix \ref{sec:tsnebert} for XLM and XLM-R plots, which exhibit similar trends to LASER and M-BERT respectively). Initially, LASER and XLM also appear to create a continuous language space by representing language relationships in terms of geometric distance between subspaces.  However, these initial subspaces become increasingly more clustered throughout the layers, thereby creating a common, shared, interlingual space in the higher layers.  Consequently, there appears to be a connection between the loss of typological information and the creation of more language-agnostic representations. Universality of LASER and XLM seems to come at the cost of their ability to retain language-specific information.

It should be noted that all encoders at some point cluster languages by family; however, M-BERT and XLM-R recover from this (at layer 10) by projecting languages back to separate subspaces. Moreover, XLM-R does not appear to organize its space differently from M-BERT and only improves on the performance patterns also seen in M-BERT. This indicates that XLM-R simply refines the mechanism deployed by M-BERT. 

%Lastly, LASER and XLM have significantly more trouble with Bulgarian compared to M-BERT and XLM-R. It is worth noting that this language has shown difficult behaviour in the past, where the authors ascribed this to fact that it is an analytic language while belonging to the Slavic family \citep{qian2016investigating}. Given that M-BERT and XLM-R improve considerably over this, it would be interesting to investigate whether typological properties are mainly encoded in the relative positioning of their language-specific subspaces in the future. This could help build intuition on how to improve models on languages such as Bulgarian when using cross-lingual training tasks.

\paragraph{Pretraining objectives}
LASER and XLM retain typological properties in higher layers to a lesser extent. Given that higher layers of a model are more tuned towards the pretraining objective, we speculate that this effect can be ascribed to their differences in the type of pretraining: LASER and XLM are trained with a cross-lingual objective vs. M-BERT and XLM-R trained on monolingual tasks only. In MT, the encoder needs to capture semantic meaning while the decoder is responsible for reconstructing that meaning in a target language. While the decoder might benefit from typological information about the target language, the encoder has no incentive from the decoder to capture such properties of the source language. Similarly, in TLM, the model can leverage information from both languages and is explicitly stimulated to align patterns from them. On the contrary, for monolingual tasks, the model must know which language it is encoding to succeed (e.g. to avoid predicting a French word for a Spanish sentence during MLM). This objective provides the model with a better incentive to retain typological properties in higher layers, as useful information can be leveraged from them to complete the tasks. Hence, cross-lingual objectives appear more suitable for training language-agnostic models. Moreover, it might not be reasonable to expect M-BERT and XLM-R to yield language-neutral representations, as their pretraining objectives do not stimulate them to learn an interlingua. This, in turn, poses challenges in zero-shot transfer on distant languages \citep{pires2019multilingual} and in resource-lean scenarios \citep{lauscher2020zero}. %\todo{Added last sentence}

%\todo{The below paragraph is a bit unclear. So it needs to be clarified and shortened, or dropped}
%The question is, however, what the aim of the model is, as the more language-agnostic models, e.g. LASER, tend to perform worse on downstream NLP tasks\todo{Is this true for XLM as well?}. We would argue that for most tasks, the integration of language-specific properties is in fact beneficial. Moreover, it is common practice to evaluate models on the same type of tasks. We would, however, argue that this leads to a skewed comparison as LASER and XLM improve on M-BERT and XLM-R with respect to their language independence. Thus, it would be interesting to compare these models on tasks in which language-agnostic representations are preferable\todo{for instance?}. 

\section{Conclusion}
In this work, we proposed methods for probing multilingual sentence encoders to investigate a wide range of typological properties. We found that all encoders are capable of capturing some typological properties related to word order, pronouns and negation. However, M-BERT and XLM-R generally outperform LASER and XLM, capturing variation along a wider range of linguistic properties. M-BERT and XLM-R's superiority is particularly evident for features pertaining to lexical properties. Moreover, we found that these properties are localizable to the lower layers of LASER and XLM, while in M-BERT and XLM-R they are encoded in all layers. We hypothesize that these differences can be ascribed to the models' pretraining tasks. We found a correspondence between the language independence of models, induced during cross-lingual training, and a loss of typological information, indicating that universality comes at the cost of language-specific information.

%In the future, it would be interesting to extend this work to a larger set of languages and models to investigate how well these results will generalize. Since our analysis shows that failure of the classifier for particular features and feature values is often related to specific languages, our reported performances are specific to the language set used in this work.
%\bibliography{emnlp2020}
%\bibliographystyle{acl_natbib}

\input{emnlp2020.bbl}
\onecolumn
\appendix

\section{Languages supported by XLM} \label{app:xlm-langs}
Bulgarian (bul), French (fra), Spanish (spa), German (deu), Greek (ell), Russian (rus), Turkish (tur), Arabic (ara), Vietnamese (vie), Thai (tha), Chinese (zho), Hindi (hin), Swahili (swa), Swedish (swe) and Urdu (urd). 
\\
\\
\noindent These languages are typologically diverse and cover all feature values used in our tasks. Thus, while the model might not have been trained on all languages used for probing, we made sure that the model was trained on languages that contain all values we probe for. Note that all other encoders support 93 (or more) languages, including all languages used in this work.

\vspace{1cm}
\section{Results for XLM and XLM-R}
\label{sec:xlm-resutls}
\begin{figure}[h!]
    \centering
    \begin{subfigure}{.45\textwidth}
    \includegraphics[width=\linewidth]{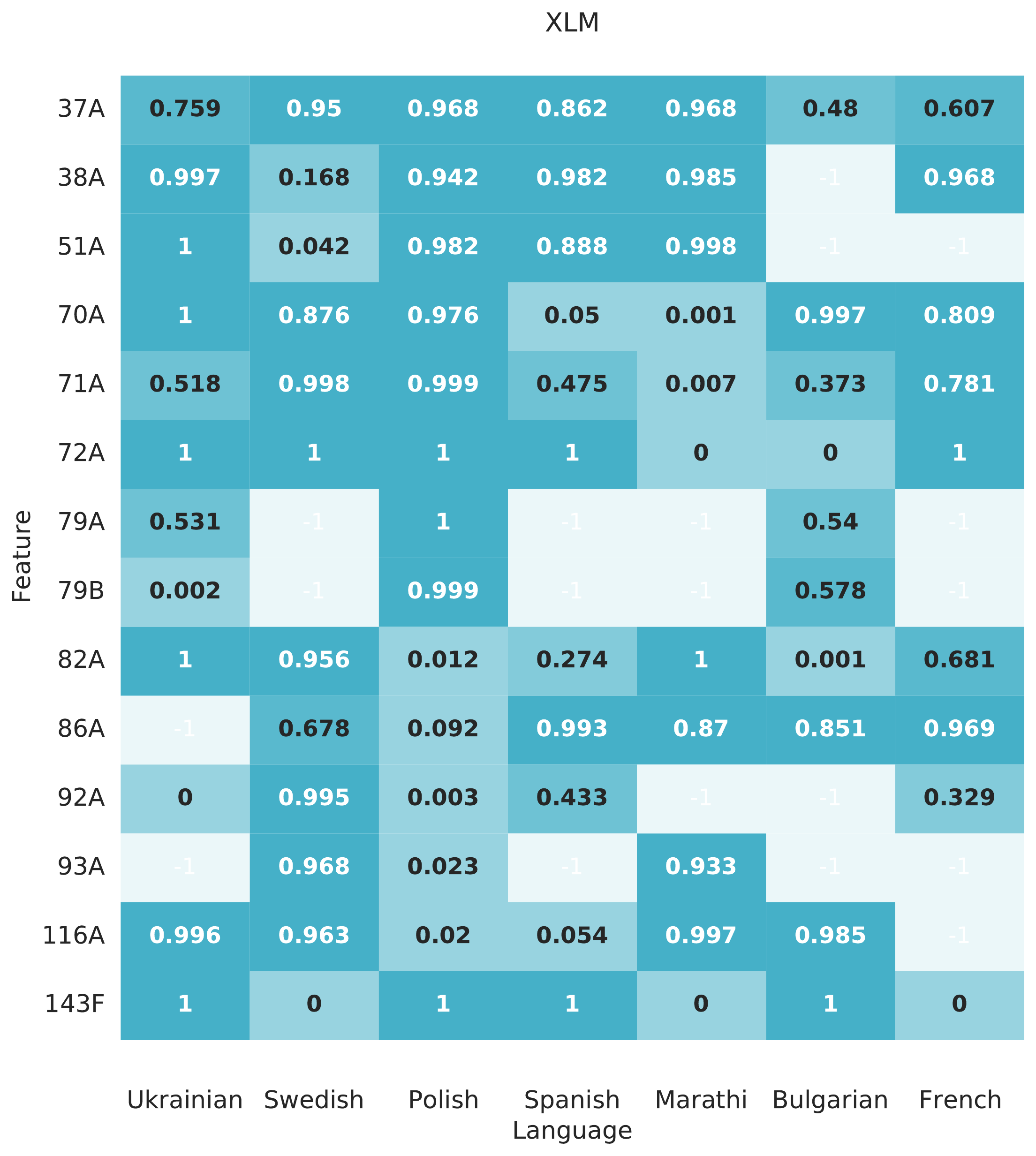}
    \end{subfigure}
    \begin{subfigure}{.45\textwidth}
     \centering
    \includegraphics[width=\linewidth]{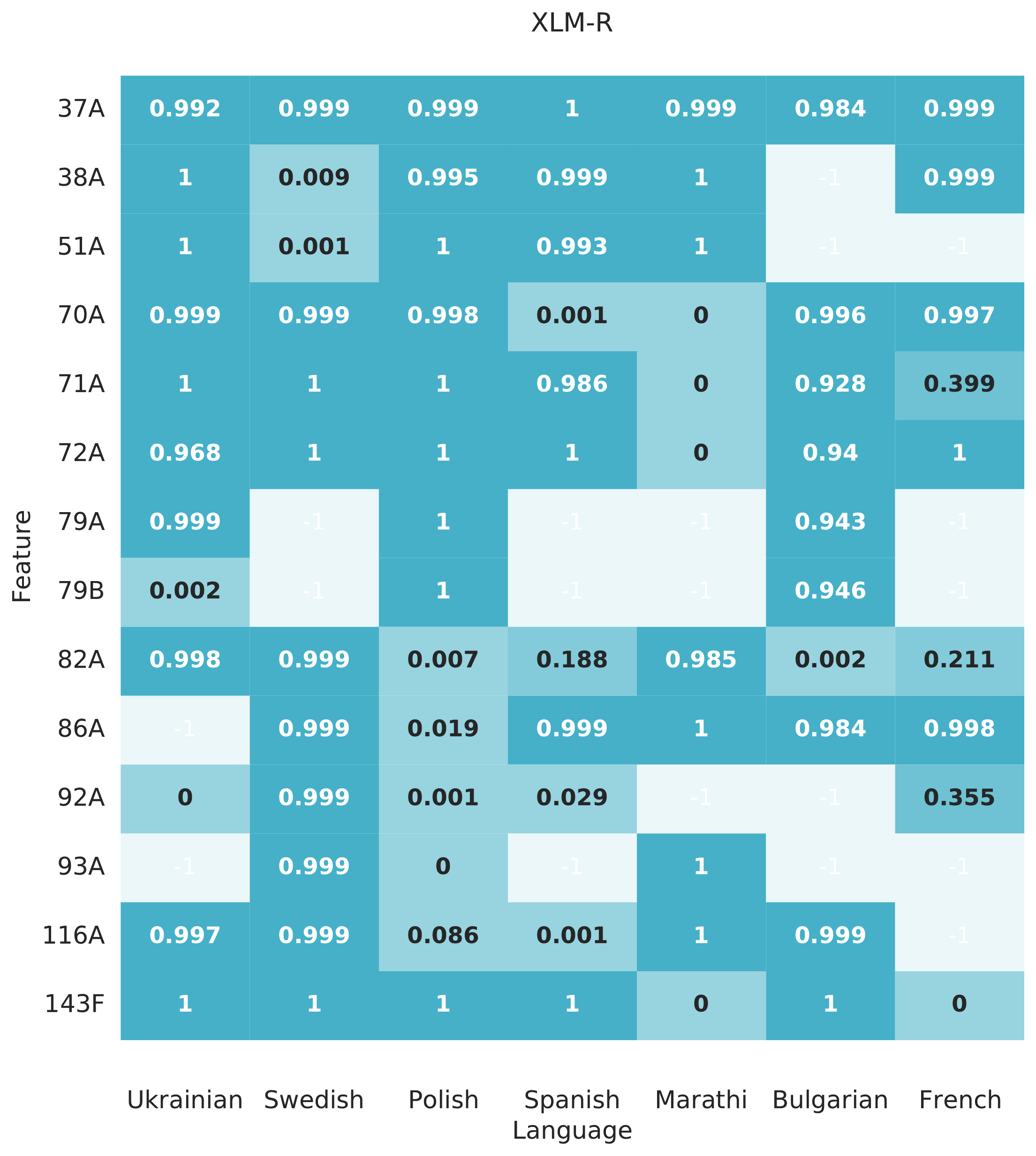}
    \end{subfigure}
    \caption{Performance (in \% accuracy) from XLM and XLM-R broken down per language. We see that XLM and XLM-R exhibit similar patterns to LASER and M-BERT. In particular, XLM-R seems to obtain its performance gain mainly from improving on languages for which M-BERT already performed well.}
\end{figure}

\begin{figure}[h!]
    \centering
    \begin{subfigure}{.45\textwidth}
    \includegraphics[width=\linewidth]{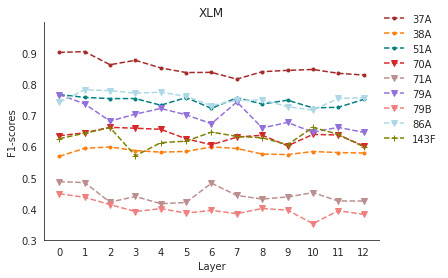}
    \end{subfigure}
    \begin{subfigure}{.45\textwidth}
     \centering
    \includegraphics[width=\linewidth]{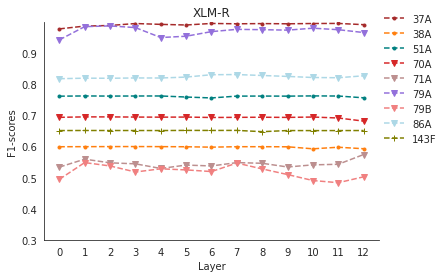}
    \end{subfigure}
    \caption{Macro-averaged-F1 scores for the representations retrieved from the different layer activations of XLM and XLM-R, where layer 0 corresponds to the non-contextualized token embeddings.}
\end{figure}
\newpage

\section{Reproducibility details}
\paragraph{Links to source code and data}
The following links can be used to download the pretrained models that we study in this work:
\begin{itemize}
    \item LASER: \href{https://dl.fbaipublicfiles.com/laser/models/bilstm.93langs.2018-12-26.pt}{BiLSTM.93langs.2018-12-2}
    \item M-BERT: \href{https://storage.googleapis.com/bert_models/2018_11_23/multi_cased_L-12_H-768_A-12.zip}{Bert-Base, multilingual cased version }
    \item XLM: \href{https://dl.fbaipublicfiles.com/XLM/mlm_tlm_xnli15_1024.pth}{xlm mlm-tlm-xnli15 }
    \item XLM-R: \href{https://dl.fbaipublicfiles.com/fairseq/models/xlmr.base.v0.tar.gz}{xlm-r.base.v0}
\end{itemize}
\noindent For the Transformers we relied on the implementations from \href{https://github.com/huggingface/transformers}{HuggingFace}, and for LASER we consulted the publicly available source code on their \href{https://github.com/facebookresearch/LASER}{GitHub repository}. Furthermore, sentences for all languages can be downloaded from the \href{https://tatoeba.org/eng/downloads}{Tatoeba website}, and to extract typological information from WALS we used the \href{https://pypi.org/project/lingtypology/}{LingTypology API}.

\paragraph{Number of model parameters}
\begin{table}[h]
    \begin{center}
     \begin{tabular}{c|cccccccc}
        \toprule 
        Model  & tokenization & L & dim & H & params & V & task & languages \\
        \midrule
        LASER & BPE & 5 & 1024 & - & 52M& 50K & MT & 93 \\
        M-BERT& WordPiece & 12 & 768 & 12 & 172M & 110K & MLM+NSP & 104\\
        XLM & BPE & 12 & 1024 & 8 & 250M & 95K & MLM+TLM & 15\\
        XLM-R & SentencePiece & 12 & 768 & 12 & 270M& 250K & MLM & 100\\
        \bottomrule
      \end{tabular}
      \caption{Summary statistics of the model architectures: tokenization method, number of layers $L$, dimensionality of sentence representations $dim$, number of attention heads $H$, number of model parameters, vocabulary size $V$ and pretraining tasks used.} 
     \label{featurespecs}
    \end{center}
\end{table}

The probing classifier has a varying number of parameters depending on the dimensionality  of the sentence representations $dim$ and the number of class labels in the task $o_n$: $params = (dim \times 100) + (100 \times o_n) + 100 + o_n$. In the scalar mixing weights experiments, another $L$ + 1 weights are added to this. See Table \ref{featurespecs} for the number of parameters in each multilingual encoder. 

\paragraph{Data set size} The number of sentences in the data sets depend on the number of language pairs $n$ included in the task. For each language we have 10K sentences, thus given $n$ language pairs we use $n \times 10K$ sentences for training. We hold out $10\%$ of our test set for validation ($ n \times 1000$) and use the remaining $n \times 9K$ sentences for testing.

\paragraph{Evaluation metric} We report results using \href{https://scikit-learn.org/stable/modules/model_evaluation.html#precision-recall-f-measure-metrics}{macro-averaged-F1 scores} as our tasks contain class imbalances, where often only a few languages are annotated with a rare class label. Instead of smoothing these class imbalances out, we assign all classes with an equal weight as we are especially interested in these minority class predictions. Thus, this is a stricter metric for our tasks than micro-averaged-F1 scores, where majority class voting as a baseline could obtain a much higher performance on most tasks.

\paragraph{Computing infrastructure} The top-layer probing experiments were run using a 2.7 GHz Intel Core i7 CPU. The other experiments required more memory and were run on the Lisa cluster, maintained by SURFsara, using a 2.10 GHz Intel Xeon Silver 4110 CPU.

\newpage

\section{Learned mixing weights}
\label{sec:scalarweights}
\vspace{1cm}
\begin{figure}[H]
    \includegraphics[width=\linewidth]{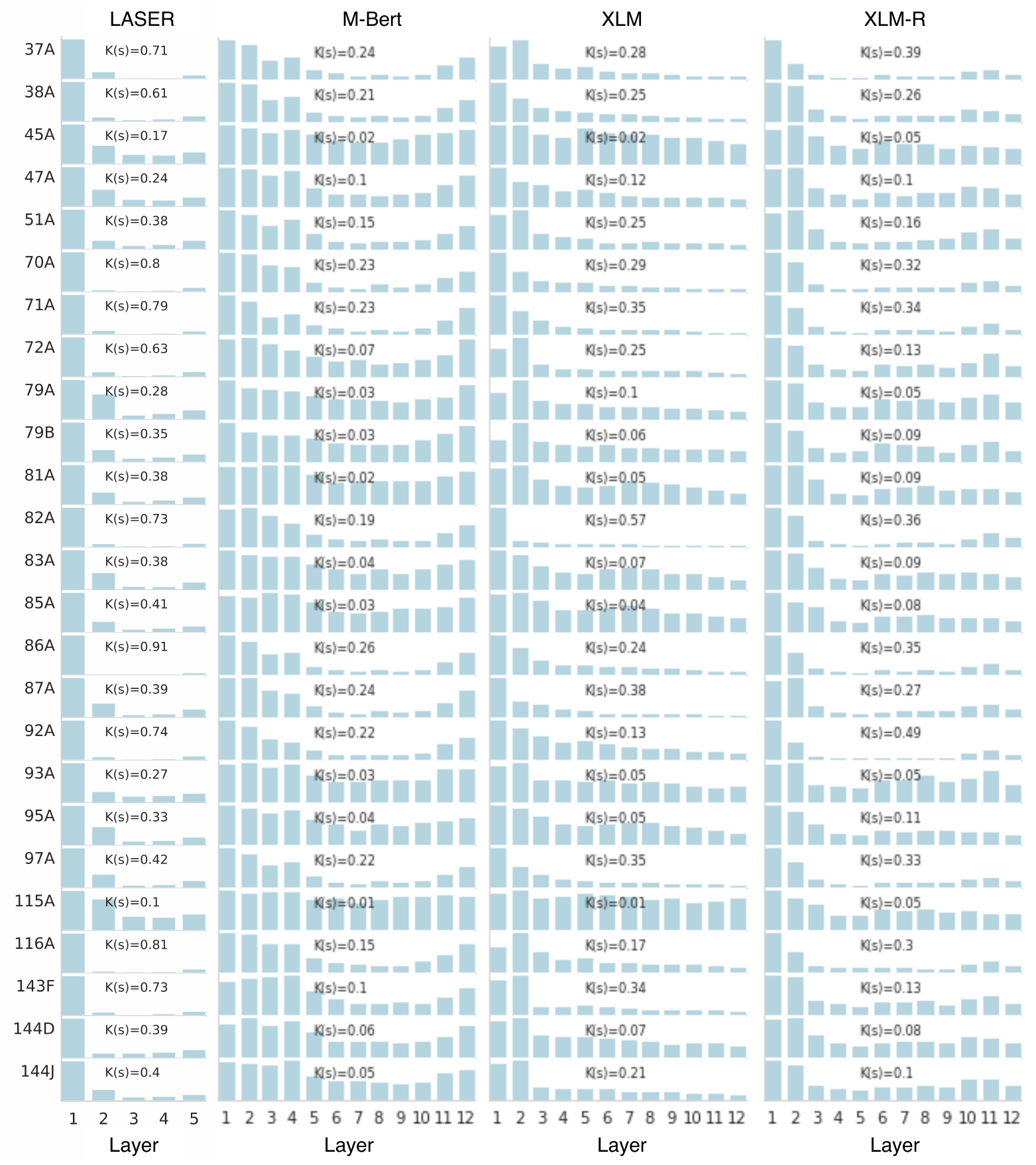}
    \caption{Learned mixing weights $s_\tau$ for each encoder and the corresponding KL divergence $K(s)$ for all 25 tasks. We see that LASER and XLM exhibit the same pattern where higher layers become less important. In M-BERT and XLM-R on the other hand, layers from $\pm$ 10 and up seem to regain importance again.}
\end{figure}
\newpage

\section{t-SNE plots per layer}
\label{sec:tsnebert}
\begin{figure}[H]
    \centering
    \includegraphics[width=\linewidth]{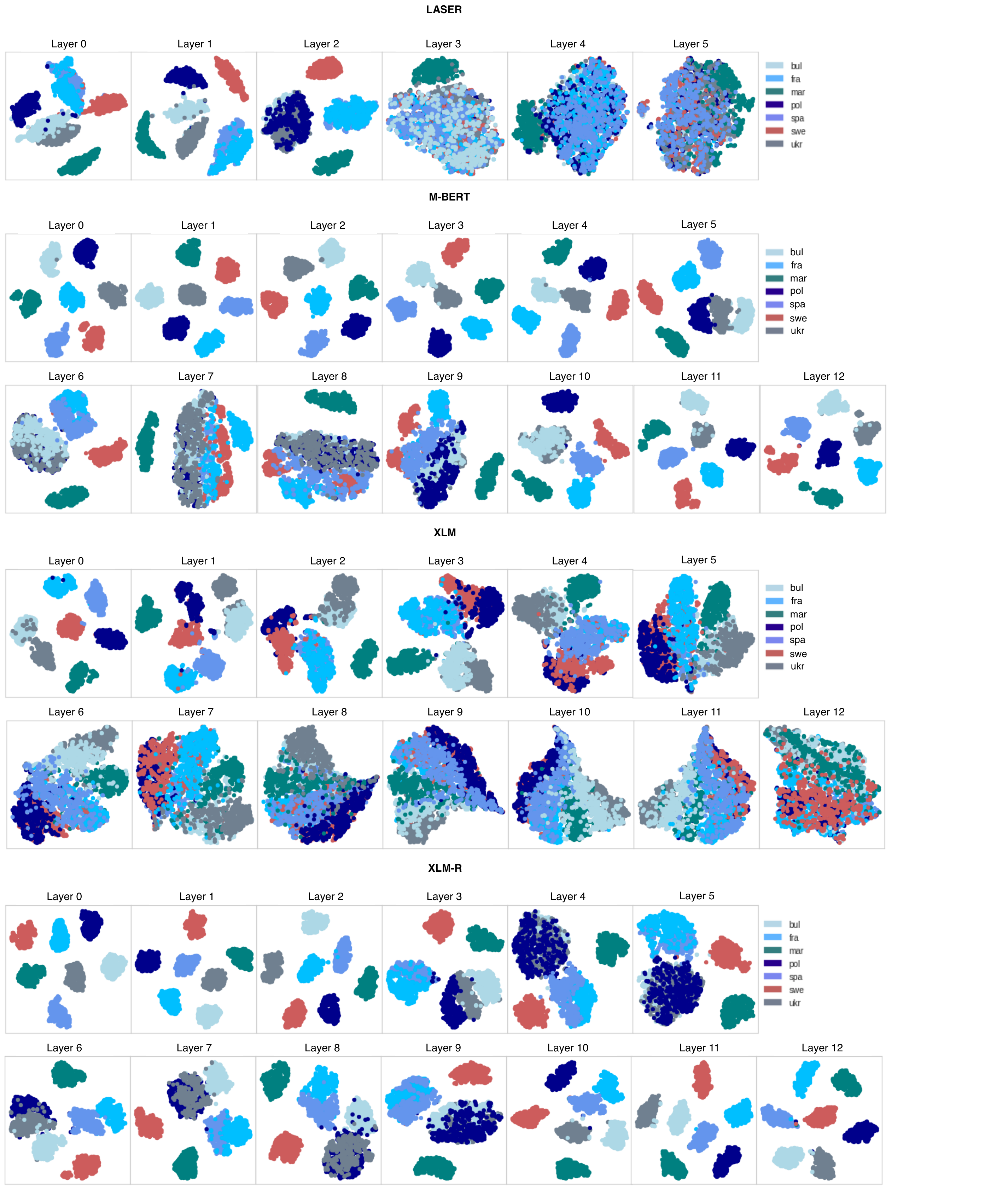}
    \caption{t-SNE visualizations of the sentence representations retrieved from the different layers of LASER, M-BERT, XLM and XLM-R, where layer 0 corresponds to the non-contextualized token embeddings (made using PCA with k = 10). Whereas LASER and XLM project all languages to a shared space in their last layers, M-BERT and XLM-R project the representations back to language-specific subspaces. Note that a similar trend is observed when only plotting the representations for the languages that XLM is trained on.}
\end{figure}
\newpage

\section{XLM generalization to unseen languages}
\label{app:generalization}
\begin{table}[H]
\centering
\begin{tabular}{ccccc}
\toprule
WALS code &   LASER non-XNLI &    XLM non-XNLI \\
\midrule
     37A &   0.305003 &   0.315463 \\
     38A &   0.325383 &   0.329041 \\
     45A &   0.498301 &   0.494767 \\
     47A &   0.481012 &   0.498201 \\
     51A &   0.498408 &   0.498352 \\
\midrule
     70A &   0.257501 &   0.266342 \\
     71A &   0.221012 &   0.209911 \\
     72A &   0.386056 &   0.400814 \\
     79A &   0.475045 &   0.433473 \\
     79B &   0.198133 &   0.221936 \\
\midrule
     81A &   0.991734 &  0.963265 \\
     82A &   0.399975 &  0.411855 \\
     83A &   0.991061 &  0.944318 \\
     85A &   0.991632 & 0.970614 \\
     86A &   0.360055 &  0.363591 \\
     87A &   0.495345 &  1.000000 \\
     92A &   0.000754 &  0.000975 \\
     93A &   0.367556 &  0.340872 \\
     95A &   0.991335 &  0.985749 \\
     97A &   0.659789 &  0.655212 \\
\midrule
    115A &   0.499235 &  0.497361 \\
    116A &   0.426288 &  0.400575 \\
\midrule
    143F &   0.400558 &  0.400814 \\
    144D &  0.499834 &   0.499734 \\
    144J &    0.499083 &   1.000000 \\
\bottomrule
\end{tabular}
\caption{macro-averaged-F1 scores for LASER and XLM computed separately over the set of XNLI languages that are not supported by XLM (Ukrainian, Polish and Marathi) (non-XNLI). Note that LASER was trained on all languages and is thus used as a comparison to the scores obtained by XLM. We see that in general, XLM obtains similar, and sometimes better, scores compared to LASER, despite not having been trained on the languages. }
\end{table}

\end{document}

%% file: table_variation.tex
\setlength{\tabcolsep}{2.6pt}

\begin{table}[t!]
\small
\begin{tabular}{l|c|c|c|c|c|c|c|c|c|c|c|c|c|c}
\midrule
&\multicolumn{7}{c|}{Train}&\multicolumn{7}{c}{Test}\\
\hline
\T & ru & da &cs & pt & hi  & mk& it& uk& sv & pl & es & mr & bg & fr \\
\midrule
37A & \cellcolor{fandango!75}&  \cellcolor{yellowish}&\cellcolor{fandango!75} & \cellcolor{teal}&\cellcolor{fandango!75} &\cellcolor{teal} &\cellcolor{teal} & \cellcolor{fandango!75}&\cellcolor{yellowish} &\cellcolor{fandango!75} & \cellcolor{teal}&\cellcolor{fandango!75} & \cellcolor{teal}&\cellcolor{teal}     \\
38A  &\cellcolor{fandango!75}&  \cellcolor{yellowish} &\cellcolor{fandango!75} & \cellcolor{teal}& \cellcolor{fandango!75}&\cellcolor{White!50} &\cellcolor{teal} &\cellcolor{fandango!75} &\cellcolor{teal} &\cellcolor{fandango!75} &\cellcolor{teal} &\cellcolor{fandango!75} & \cellcolor{White!50}& \cellcolor{teal}    \\
51A &\cellcolor{teal} &\cellcolor{teal}  & \cellcolor{teal}&\cellcolor{fandango!75} &\cellcolor{teal} &\cellcolor{White!50} & \cellcolor{White!50}&\cellcolor{teal} &\cellcolor{fandango!75} &\cellcolor{teal} &\cellcolor{fandango!75} &\cellcolor{teal} &\cellcolor{White!50} & \cellcolor{White!50}    \\
70A &\cellcolor{teal} & \cellcolor{fandango!75} & \cellcolor{teal} &\cellcolor{yellowish} & \cellcolor{teal} &\cellcolor{teal}  &\cellcolor{yellowish} &\cellcolor{teal}  &\cellcolor{fandango!75} &\cellcolor{teal}  &\cellcolor{teal}  &\cellcolor{fandango!75} &\cellcolor{teal}  & \cellcolor{yellowish}    \\
71A & \cellcolor{teal}&\cellcolor{teal}  &\cellcolor{teal} &\cellcolor{fandango!75} & \cellcolor{greenish}& \cellcolor{yellowish}&\cellcolor{fandango!75} &\cellcolor{teal} &\cellcolor{teal} &\cellcolor{teal} &\cellcolor{fandango!75} &\cellcolor{yellowish} &\cellcolor{yellowish} & \cellcolor{teal}    \\
72A & \cellcolor{teal}&\cellcolor{teal}  &\cellcolor{teal} &\cellcolor{teal} & \cellcolor{teal}& \cellcolor{fandango!75}&\cellcolor{teal} &\cellcolor{teal} &\cellcolor{teal} &\cellcolor{teal} &\cellcolor{teal} &\cellcolor{fandango!75} &\cellcolor{fandango!75} & \cellcolor{teal}   \\
79A & \cellcolor{teal}&\cellcolor{White!50} &\cellcolor{teal} &\cellcolor{White!50} &\cellcolor{White!50}&\cellcolor{fandango!75} &\cellcolor{White!50} & \cellcolor{teal} &\cellcolor{White!50} &  \cellcolor{teal} &\cellcolor{White!50}&\cellcolor{White!50} & \cellcolor{fandango!75} &\cellcolor{White!50}     \\
79B  &\cellcolor{fandango!75}&\cellcolor{White!50} &\cellcolor{teal} &\cellcolor{White!50} &\cellcolor{White!50}&\cellcolor{yellowish} &\cellcolor{White!50} & \cellcolor{teal} &\cellcolor{White!50} &  \cellcolor{teal} &\cellcolor{White!50}&\cellcolor{White!50} & \cellcolor{yellowish} &\cellcolor{White!50}     \\
82A &  \cellcolor{teal}&\cellcolor{teal}  &\cellcolor{teal} &\cellcolor{teal} & \cellcolor{teal}& \cellcolor{teal}&\cellcolor{fandango!75} &\cellcolor{teal} &\cellcolor{teal} &\cellcolor{fandango!75} &\cellcolor{fandango!75} &\cellcolor{teal} &\cellcolor{fandango!75} & \cellcolor{teal}   \\
86A &  \cellcolor{White!50}&\cellcolor{fandango!75}  &\cellcolor{yellowish} &\cellcolor{teal} & \cellcolor{fandango!75}& \cellcolor{yellowish}&\cellcolor{teal} &\cellcolor{White!50} &\cellcolor{fandango!75} &\cellcolor{teal} &\cellcolor{teal} &\cellcolor{fandango!75} &\cellcolor{yellowish} & \cellcolor{teal}   \\
92A &\cellcolor{yellowish} &\cellcolor{teal}  &\cellcolor{teal} & \cellcolor{fandango!75}& \cellcolor{White!50}& \cellcolor{White!50}&\cellcolor{teal} &\cellcolor{fandango!75} &\cellcolor{teal} &\cellcolor{fandango!75} &\cellcolor{teal} & \cellcolor{White!50}&\cellcolor{White!50} &\cellcolor{fandango!75}     \\
93A & \cellcolor{teal}& \cellcolor{teal} &\cellcolor{fandango!75} &\cellcolor{White!50}  &\cellcolor{fandango!75} &\cellcolor{White!50}  &\cellcolor{White!50}  &\cellcolor{teal} &\cellcolor{teal} &\cellcolor{teal} &\cellcolor{White!50}  &\cellcolor{fandango!75} &\cellcolor{White!50}  &\cellcolor{White!50}      \\
116A &\cellcolor{teal} &\cellcolor{fandango!75}& \cellcolor{fandango!75} &\cellcolor{teal} &\cellcolor{teal} &\cellcolor{teal} &\cellcolor{White!50} &\cellcolor{teal} &\cellcolor{fandango!75}  & \cellcolor{teal}&\cellcolor{fandango!75} &\cellcolor{teal} &\cellcolor{teal} &  \cellcolor{White!50}   \\
143F &\cellcolor{teal} & \cellcolor{fandango!75} &\cellcolor{teal} &\cellcolor{teal} &\cellcolor{teal} & \cellcolor{teal}& \cellcolor{teal}& \cellcolor{teal}&\cellcolor{fandango!75} & \cellcolor{teal}&\cellcolor{teal}  &\cellcolor{fandango!75} &\cellcolor{teal} &\cellcolor{fandango!75}     \\
\midrule
\end{tabular}
\vspace{-0.2cm}
\caption{Color coding of the typological diversity of the train and test languages w.r.t. different features. Languages with the same color have the same feature value for that task (excluded languages are left blank).}
\label{table:variation}
\end{table}